\documentclass{article}
\usepackage{spconf}
\usepackage{amsmath}
\usepackage{fontspec}
\usepackage{unicode-math}
\usepackage{graphicx,booktabs,array,xcolor,cite,pifont}
\usepackage{tikz}

\usetikzlibrary{arrows.meta,shadings}
\usepackage{enumitem}
\setlist{nosep, leftmargin=*, labelsep=4pt}
\usepackage[hidelinks]{hyperref}
\hypersetup{pdftitle={When Cell Crops Share a Parent: Measuring and Isolating Acquisition Leakage in Cervical Cytology Benchmarks},
  pdfauthor={Snehasish Ghosh, Soham Mitra, Antareep Majumder, Nibaran Das},
  pdfkeywords={cervical cytology, data leakage, evaluation protocol, shortcut learning, benchmark validity, preregistration}}
\makeatletter
\def\thebibliography#1{\section{References}\small\list
 {[\arabic{enumi}]}{\settowidth\labelwidth{[#1]}\leftmargin\labelwidth
 \advance\leftmargin\labelsep\usecounter{enumi}%
 \setlength{\itemsep}{0.8pt}\setlength{\parsep}{0pt}\setlength{\topsep}{2pt}}
 \def\newblock{\hskip .11em plus .33em minus .07em}
 \sloppy\clubpenalty4000\widowpenalty4000\sfcode`\.=1000\relax}
\makeatother

\makeatletter
\long\def\@makecaption#1#2{\vskip 3pt\small
  \setbox\@tempboxa\hbox{#1. #2}%
  \ifdim\wd\@tempboxa>\hsize #1. #2\par\else\hbox to\hsize{\hfil\box\@tempboxa\hfil}\fi\vskip 1pt}
\makeatother
\definecolor{ink}{HTML}{0B0B0B}   \definecolor{ink2}{HTML}{52514E}
\definecolor{rule}{HTML}{A9A8A2}  \definecolor{other}{HTML}{E4E3DE}
\definecolor{leak}{HTML}{EB6834}  \definecolor{clean}{HTML}{2A78D6}
\definecolor{fieldA}{HTML}{EBCFE3} \definecolor{fieldB}{HTML}{CFDDF0}
\definecolor{fieldC}{HTML}{DDE9C9} \definecolor{fieldD}{HTML}{F3E1C8}
\definecolor{cyto}{HTML}{FFFFFF}  \definecolor{nuc}{HTML}{5A3F7A}
\definecolor{featfill}{HTML}{7FA9DE} \definecolor{heat}{HTML}{F2A33A}

\newcommand{\yes}{\textcolor{clean}{\ding{51}}}
\newcommand{\no}{\textcolor{rule}{\ding{55}}}
\newcommand{\pp}{\,pp}

\newcommand{\CamBgAttribution}{21.8}

\newcommand{\CamLeakEst}{\textminus{}0.09}
\newcommand{\CamLeakHi}{0.39}
\newcommand{\CamLeakLo}{\textminus{}0.57}

\newcommand{\CamSpearman}{0.08}
\newcommand{\CarrierColour}{1.71}

\newcommand{\CarrierStandard}{1.47}

\newcommand{\CricPOneAcc}{90.48}
\newcommand{\CricPOneBacc}{78.53}
\newcommand{\CricPTwoAcc}{78.40}
\newcommand{\CricPTwoBacc}{58.89}

\newcommand{\CricPrimeEst}{12.11}

\newcommand{\DivEst}{0.00}
\newcommand{\DivHi}{0.64}
\newcommand{\DivLo}{\textminus{}0.65}

\newcommand{\FeightCells}{15}
\newcommand{\FeightEst}{0.000}
\newcommand{\FeightHi}{0.005}
\newcommand{\FeightLo}{\textminus{}0.005}

\newcommand{\FeightPholm}{1.00}

\newcommand{\FfiveEst}{\textminus{}0.01}
\newcommand{\FfiveHi}{0.07}
\newcommand{\FfiveLo}{\textminus{}0.10}

\newcommand{\FfivePholm}{1.00}

\newcommand{\FfourEst}{\textminus{}0.23}
\newcommand{\FfourHi}{0.35}
\newcommand{\FfourLo}{\textminus{}0.81}

\newcommand{\FfourPholm}{1.00}

\newcommand{\FoneEst}{\textminus{}4.29}
\newcommand{\FoneHi}{\textminus{}1.37}
\newcommand{\FoneLo}{\textminus{}8.60}

\newcommand{\FonePholm}{0.001}

\newcommand{\FsevenEst}{0.000}
\newcommand{\FsevenHi}{0.039}
\newcommand{\FsevenLo}{\textminus{}0.038}

\newcommand{\FsevenPholm}{1.00}

\newcommand{\FsixEst}{0.02}
\newcommand{\FsixHi}{0.44}
\newcommand{\FsixLo}{\textminus{}0.41}

\newcommand{\FsixPholm}{1.00}

\newcommand{\FthreeEst}{1.86}
\newcommand{\FthreeHi}{2.30}
\newcommand{\FthreeLo}{1.42}

\newcommand{\FthreePholm}{$<$0.001}

\newcommand{\FtwoEst}{1.27}
\newcommand{\FtwoHi}{2.16}
\newcommand{\FtwoLo}{0.38}

\newcommand{\FtwoPholm}{0.03}

\newcommand{\GrlConvRstar}{55.1}
\newcommand{\GrlConvRzero}{29.5}

\newcommand{\HfiveEst}{0.06}
\newcommand{\HfiveHi}{0.12}
\newcommand{\HfiveLo}{\textminus{}0.01}

\newcommand{\HfivePholm}{0.16}

\newcommand{\HfourEst}{0.42}
\newcommand{\HfourHi}{1.18}
\newcommand{\HfourLo}{\textminus{}0.35}

\newcommand{\HfourPholm}{0.23}

\newcommand{\HoneArchMax}{1.44}
\newcommand{\HoneArchMin}{0.66}

\newcommand{\HoneBootHi}{1.39}
\newcommand{\HoneBootLo}{0.59}
\newcommand{\HoneBootP}{$<$0.001}

\newcommand{\HoneEst}{0.97}

\newcommand{\HoneKRp}{$<$0.001}

\newcommand{\HthreeEst}{1.73}
\newcommand{\HthreeHi}{2.51}
\newcommand{\HthreeLo}{0.95}

\newcommand{\HthreePholm}{0.02}
\newcommand{\HthreeRepEst}{1.47}

\newcommand{\HtwoEst}{12.08}
\newcommand{\HtwoHi}{12.99}

\newcommand{\HtwoLo}{11.16}

\newcommand{\HtwoPholm}{$<$0.001}

\newcommand{\InteractionEst}{\textminus{}0.20}

\newcommand{\InteractionP}{0.60}

\newcommand{\KendallTau}{0.82}
\newcommand{\LeaceAccDrop}{0.17}
\newcommand{\LeaceGapTop}{0.95}
\newcommand{\LeaceGapZero}{0.93}
\newcommand{\LeaceRtop}{5.4}
\newcommand{\LeaceRzero}{25.1}

\newcommand{\LeakUnderBg}{1.17}
\newcommand{\LeakUnderFlat}{1.30}
\newcommand{\LeakUnderOriginal}{1.72}
\newcommand{\LeakUnderStain}{0.46}

\newcommand{\OverlapNo}{1.68}

\newcommand{\OverlapYes}{1.78}

\newcommand{\PowerChosen}{0.977}

\newcommand{\PtwoPthreeEst}{0.03}
\newcommand{\PtwoPthreeHi}{0.56}

\newcommand{\PtwoPthreeLo}{\textminus{}0.49}

\newcommand{\RFloor}{1.2}

\newcommand{\SipPOneAcc}{97.14}
\newcommand{\SipPTwoAcc}{96.18}

\newcommand{\RecFollowTasks}{1,580}
\newcommand{\RecFollowTrainRuns}{940}

\newcommand{\RecOverlapCric}{71.5}
\newcommand{\RecOverlapCx}{34.0}
\newcommand{\RecOverlapSip}{52.6}

\newcommand{\RecStudyTwoRuns}{1,341}

\newcommand{\LeaceAboveFloorCut}{82}

\title{When cell crops share a parent: measuring and isolating acquisition leakage in cervical cytology benchmarks}

\name{Snehasish Ghosh$^{1,*}$ \qquad Soham Mitra$^{2,*}$ \qquad Antareep Majumdar$^{3}$ \qquad KC Santosh$^{4}$ \qquad
Nibaran Das$^{2}$}

\address{\normalsize
$^{1}$Independent Researcher, formerly affiliated with the Indian Institute of Science, Bengaluru, India\\
\normalsize $^{2}$Department of Computer Science and Engineering, Jadavpur University, Kolkata, India\\
\normalsize $^{3}$Qusbi Infotech Private Limited, Kolkata, India,
\normalsize $^{4}$Oregon Institute of Technology, USA\\
\normalsize $^{*}$These authors contributed equally to this work.}

\begin{document}
\maketitle

\begin{abstract}
Public cervical cytology benchmarks release many cell crops cut from far fewer microscope fields or slides and are often split at the crop level, so test cells share acquisition parents with training cells. In a preregistered study, crop-level splitting raised SIPaKMeD accuracy by \HoneEst\ percentage points across twelve architectures and twenty seeds (95\% CI \HoneBootLo--\HoneBootHi), and CRIC accuracy by \HtwoEst\ points (\HtwoLo--\HtwoHi). On SIPaKMeD, a factorial that fixes training crop and parent counts attributes the gain to sibling exposure (\HthreeEst\ points, \HthreeLo--\HthreeHi), not to the extra parent diversity of crop-level splits (\DivEst). Parent identity is decodable at about 100 times chance, yet erasing \LeaceAboveFloorCut\% of its above-floor linear recoverability left the gap unchanged, and Grad-CAM did not distinguish leaking from clean models; matching test cells' color statistics to an unseen same-class field removed most of the advantage. Decodability and saliency cannot substitute for a parent-disjoint split.
\end{abstract}

\begin{keywords}
Cervical cytology, data leakage, shortcut learning, benchmark validity, preregistration, evaluation protocol
\end{keywords}

\section{Introduction}
\label{sec:intro}

\begin{figure*}[t]
\centering
% Study overview: hierarchy and protocols (a), matched-diversity factorial (b), diagnostics (c).
% Drawn at the full text width (178 mm); \input inside a figure*. Colours are defined in main.tex.
\begin{tikzpicture}[
    x=1cm, y=1cm, font=\sffamily\footnotesize,
    lbl/.style={font=\sffamily\footnotesize, text=ink2, inner sep=1pt},
    small/.style={font=\sffamily\scriptsize, text=ink2, inner sep=1pt},
    ttl/.style={font=\sffamily\small, text=ink, anchor=west, inner sep=0pt},
    box/.style={draw=rule, rounded corners=2pt, line width=0.5pt, fill=white, inner sep=0pt},
    arr/.style={-{Stealth[length=4pt,width=3.5pt]}, line width=0.6pt, draw=ink2},
    pics/cellglyph/.style={code={
        \fill[cyto] (0,0) ellipse (0.14 and 0.10);
        \fill[nuc] (0.03,0.01) circle (0.042);
    }},
    pics/crop/.style={code={
        \filldraw[fill=#1, draw=rule, line width=0.4pt] (-0.19,-0.19) rectangle (0.19,0.19);
        \pic {cellglyph};
    }},
    pics/mini/.style={code={
        \filldraw[fill=#1, draw=rule, line width=0.35pt] (-0.1,-0.1) rectangle (0.1,0.1);
    }},
    pics/field/.style={code={
        \filldraw[fill=#1, draw=rule, rounded corners=2pt, line width=0.4pt] (0,0) rectangle (1.35,1.15);
        \foreach \x/\y in {0.36/0.84, 0.99/0.86, 0.38/0.31, 0.97/0.33} {
          \pic at (\x,\y) {cellglyph};
          \draw[draw=ink2, densely dashed, line width=0.35pt] (\x-0.19,\y-0.19) rectangle (\x+0.19,\y+0.19);
        }
    }},
  ]

% ------------------------------------------------------------------ (a) hierarchy and protocols
\node[ttl] at (0,4.45) {\textbf{(a)}\enspace Acquisition hierarchy and splits};
\pic at (0,2.55) {field=fieldA};
\pic at (0,0.75) {field=fieldB};
\node[lbl, anchor=south] at (0.675,3.72) {field A};
\node[lbl, anchor=north] at (0.675,0.72) {field B};
\node[small, align=center] at (0.675,2.23) {dashed: released\\cell crops};

\node[lbl, anchor=south] at (3.35,3.55) {train};
\node[lbl, anchor=south] at (4.985,3.55) {test};
% P1
\node[font=\sffamily\footnotesize\bfseries, text=ink] at (1.98,3.15) {P1};
\node[box, minimum width=1.8cm, minimum height=0.6cm] at (3.35,3.15) {};
\node[box, minimum width=1.03cm, minimum height=0.6cm, draw=leak, line width=0.8pt] at (4.985,3.15) {};
\pic at (2.72,3.15) {crop=fieldA}; \pic at (3.14,3.15) {crop=fieldA};
\pic at (3.56,3.15) {crop=fieldB}; \pic at (3.98,3.15) {crop=fieldB};
\pic at (4.74,3.15) {crop=fieldA}; \pic at (5.23,3.15) {crop=fieldB};
\draw[leak, line width=0.8pt, -{Stealth[length=4pt,width=3.5pt]}]
  (3.14,2.93) .. controls (3.5,2.45) and (4.4,2.45) .. (4.74,2.93);
\node[small, text=leak, anchor=north] at (3.94,2.52) {crop-level: siblings cross the split};
% P2
\node[font=\sffamily\footnotesize\bfseries, text=ink] at (1.98,1.45) {P2};
\node[box, minimum width=1.8cm, minimum height=0.6cm] at (3.35,1.45) {};
\node[box, minimum width=1.03cm, minimum height=0.6cm, draw=clean, line width=0.8pt] at (4.985,1.45) {};
\foreach \x in {2.72,3.14,3.56,3.98} \pic at (\x,1.45) {crop=fieldA};
\pic at (4.74,1.45) {crop=fieldB}; \pic at (5.23,1.45) {crop=fieldB};
\node[small, text=clean, anchor=north] at (3.94,1.08) {parent-disjoint: test fields unseen};

% ------------------------------------------------------------------ (b) matched-diversity factorial
\begin{scope}[xshift=6.2cm]
\node[ttl] at (0,4.45) {\textbf{(b)}\enspace Leakage vs.\ diversity, matched size};
\node[small, anchor=south] at (0.5,3.72) {fields $E$};
\foreach \i/\c in {0/fieldA, 1/fieldB, 2/fieldD} {
  \pgfmathsetmacro{\yy}{3.25-0.55*\i}
  \filldraw[fill=\c, draw=rule, line width=0.4pt] (0.04,\yy) rectangle (0.46,\yy+0.42);
  \filldraw[fill=\c, draw=rule, line width=0.4pt] (0.54,\yy) rectangle (0.96,\yy+0.42);
  \pic[scale=0.9] at (0.25,\yy+0.21) {cellglyph};
  \pic[scale=0.9] at (0.75,\yy+0.21) {cellglyph};
}
\draw[arr] (0.25,2.12) -- (0.25,1.66);
\node[box, draw=ink2, inner sep=2.5pt, font=\sffamily\scriptsize, text=ink] (T) at (0.52,1.42) {test set $T$};
\node[small, align=center, anchor=north] at (0.52,1.12) {one per seed,\\shared by all\\four cells};
\draw[arr, draw=leak] (1.0,3.46) -- (1.98,3.46);
\node[small, text=leak, anchor=south] at (1.47,3.48) {siblings};

\def\gx{2.06}
\node[lbl, anchor=south] at (\gx+0.8,3.83) {720 fields};
\node[lbl, anchor=south] at (\gx+2.52,3.83) {504 fields};
\foreach \col/\xx in {0/2.06, 1/3.78} {
  \node[box, draw=leak, line width=0.8pt, minimum width=1.6cm, minimum height=0.86cm, anchor=north west] at (\xx,3.83) {};
  \node[small, anchor=north west, text=leak, font=\sffamily\scriptsize\bfseries] at (\xx+0.06,3.78) {leak};
  \foreach \k in {0,1,2} \pic at (\xx+0.25+0.27*\k,3.22) {mini=other};
  \pic at (\xx+0.25+0.27*3,3.22) {mini=fieldA};
  \pic at (\xx+0.25+0.27*4,3.22) {mini=fieldB};
  \node[box, draw=clean, line width=0.8pt, minimum width=1.6cm, minimum height=0.86cm, anchor=north west] at (\xx,2.87) {};
  \node[small, anchor=north west, text=clean, font=\sffamily\scriptsize\bfseries] at (\xx+0.06,2.82) {clean};
  \foreach \k in {0,...,4} \pic at (\xx+0.25+0.27*\k,2.26) {mini=other};
}
\node[small, align=center, anchor=north] at (\gx+1.66,1.92)
  {1{,}100 crops in every cell; crop and\\field counts equal within a column};
\node[small, align=left, anchor=north west, text=ink] at (\gx+0.05,1.2)
  {leak $-$ clean $\Rightarrow$ leakage (H3)\\720 $-$ 504 (clean) $\Rightarrow$ diversity};
\end{scope}

% ------------------------------------------------------------------ (c) diagnostics
\begin{scope}[xshift=12.2cm]
\node[ttl] at (0,4.45) {\textbf{(c)}\enspace Is parent identity used?};
% colour transfer
\pic at (0.22,3.42) {crop=fieldA};
\draw[arr] (0.47,3.42) -- (0.93,3.42);
\pic at (1.18,3.42) {crop=fieldC};
\node[small, anchor=west, align=left] at (1.55,3.42) {colour statistics moved to an\\unseen field of the class (F2)};
% erasure
\node[draw=ink2, line width=0.4pt, minimum width=0.15cm, minimum height=0.56cm, inner sep=0, fill=featfill] (h) at (0.08,2.3) {};
\node[draw=ink2, line width=0.4pt, fill=white, font=\sffamily\scriptsize, inner sep=1.6pt] (P) at (0.6,2.3) {$P_r$};
\node[draw=ink2, line width=0.4pt, minimum width=0.15cm, minimum height=0.56cm, inner sep=0, fill=featfill!35] (h2) at (1.12,2.3) {};
\draw[arr] (h) -- (P); \draw[arr] (P) -- (h2);
\node[small, anchor=west, align=left] at (1.55,2.3) {identity erased from features,\\head refit (F5); adversaries (F7)};
% Grad-CAM
\filldraw[fill=fieldA, draw=rule, line width=0.4pt] (-0.1,0.94) rectangle (0.46,1.5);
\fill[cyto] (0.18,1.22) ellipse (0.19 and 0.14);
\fill[nuc] (0.21,1.23) circle (0.055);
\shade[inner color=heat, outer color=fieldA, opacity=0.8] (0.2,1.22) circle (0.25);
\draw[arr] (0.55,1.22) -- (0.95,1.22);
\node[small] at (1.18,1.22) {?};
\node[small, anchor=west, align=left] at (1.55,1.22) {Grad-CAM mass outside the\\cell mask, leak vs.\ clean};
\node[small, anchor=north west, align=left, text=ink] at (-0.1,0.66)
  {use is shown only if the leak advantage\\moves; decodability alone shows availability};
\end{scope}
\end{tikzpicture}
\caption{Study design. (a) A crop-level split (P1) places siblings from one acquisition parent in training and test; a parent-disjoint split (P2) does not. (b) Matched-size factorial: one test set per seed; leak and clean training sets have identical crop and field counts and differ only in the test fields' siblings, while diversity varies separately. (c) Diagnostics asking whether decodable parent identity is what the leak uses.}
\label{fig:design}
\end{figure*}
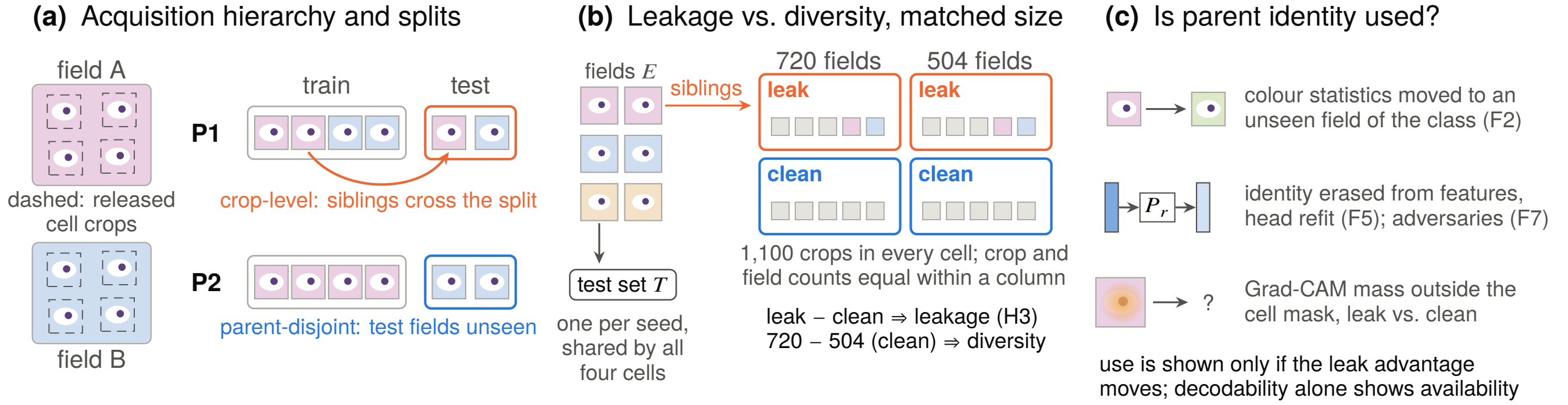

Public cervical cytology benchmarks are mostly released as cell crops cut from far fewer microscope fields or slide images~\cite{plissiti2018sipakmed,liu2022cx22,rezende2021cric}. Sibling crops share their field's stain, illumination, focus and background, and sometimes pixels. A crop-level split, the default for files that look like independent images, puts siblings on both sides of the split (Fig.~\ref{fig:design}a).

That sibling crops inflate scores is expected: placing tiles of one subject in training and validation inflated digital pathology scores by up to 41\%~\cite{bussola2021tiles}, and non-independent splits and shortcut cues mislead medical image evaluation more broadly~\cite{kapoor2023leakage,geirhos2020shortcut,varoquaux2022failures}. Benchmark artifacts have been audited in chest radiographs~\cite{jimenez2023shortcuts}, shortcut mechanisms dissected in segmentation~\cite{lin2024shortcut}, and image-level splitting shown to inflate CT accuracy~\cite{obeten2026kidney}. Contamination does not imply consequence, however: in brain-tumor MRI, removing heavily contaminated test images, without retraining, left balanced accuracy essentially unchanged~\cite{vangala2026scores}. We ask the complementary question of when acquisition siblings do change the score, which requires retraining under each protocol rather than filtering the test set.

Three questions follow. \emph{How much} does crop-level splitting change public cytology benchmarks? \emph{Is the change leakage?} A grouped split removes the test parents' siblings but, at a fixed crop budget, also leaves fewer distinct training parents (crop-level training sets hold 18\% more on SIPaKMeD), so a lower score could reflect lost diversity, a confound that split comparisons leave in place. \emph{Can common diagnostics certify an evaluation?} Decodable source identity is increasingly read as a sign of shortcut use~\cite{germani2026aggregation}, and saliency maps as evidence of where a model looks. A preregistered study and a registered follow-up (2,281 trained models) contribute:
\begin{itemize}
\item \textbf{Structure:} closed forms that predict a split's exposure from its crop-to-parent table (Sec.~\ref{sec:data}).
\item \textbf{Consequence:} one simulation-powered test per benchmark across twelve architectures (Sec.~\ref{sec:res-consequence}).
\item \textbf{Causal isolation:} a SIPaKMeD factorial that varies sibling exposure at fixed crop and parent counts (Sec.~\ref{sec:res-isolation}).
\item \textbf{Diagnostic caution:} interventions showing that the linearly decodable parent code and saliency do not track the leak (Sec.~\ref{sec:res-diagnostics}).
\end{itemize}

\section{Data and acquisition hierarchy}
\label{sec:data}

\begin{table}[t]
\caption{Audited datasets. \emph{Sib.}: test crops with a training sibling under a random stratified 80:20 crop split; \emph{Pix.}: test crops sharing source pixels with a training crop (means over five splits). Classifiers were trained only where marked; Cx22 and Herlev received the structural audit only.}
\label{tab:data}
\centering\small\setlength{\tabcolsep}{3.2pt}
\begin{tabular}{@{}llrrrrc@{}}
\toprule
Dataset & Parent & Crops & Parents & Sib.\,\% & Pix.\,\% & Models \\
\midrule
SIPaKMeD~\cite{plissiti2018sipakmed} & field & 4,049 & 966 & 92.6 & \RecOverlapSip & \yes \\
Cx22~\cite{liu2022cx22} & image & 1,320 & 514 & 85.2 & \RecOverlapCx & -- \\
CRIC~\cite{rezende2021cric} & slide & 11,534 & 400 & 99.97 & \RecOverlapCric & \yes \\
Herlev~\cite{jantzen2005pap} & cell & 917 & 917 & 0.0 & 0.0 & -- \\
\bottomrule
\end{tabular}
\end{table}

Table~\ref{tab:data} lists four public datasets. SIPaKMeD crops come from cluster fields; filename parsing and template matching independently assign all 4,049 crops to the same field. Cx22 regions come from CCEDD source images, CRIC cells from 400 slide images with nucleus coordinates, and Herlev, with one cell per image, is a negative control. A \emph{parent} is the deepest acquisition unit a release identifies; it is not a patient.

If a parent's $k$ crops are allocated independently to a test fraction $q$, a test crop has a training sibling with probability $1-q^{k-1}$, and at equal size a crop-level training set holds $\sum_j(1-q^{k_j})/\{(1-q)m\}$ times as many distinct parents as a parent-level one over $m$ parents (1.18 for SIPaKMeD, 1.25 for CRIC). These closed forms reproduce Table~\ref{tab:data}. The parent is also learnable: a ResNet-18 trained to identify 424 SIPaKMeD fields recognised held-out siblings at 25.7\% (chance 0.24\%), and at 18.7\% from cell-free backgrounds. Each SIPaKMeD field holds one class; in CRIC, slide identity carries 68\% of the class entropy.

\section{Study design}
\label{sec:design}

Everything reported comes from the preregistered Study~2 and its registered follow-up; an earlier pilot (Study~1, six architectures, five seeds) served only the power analysis. Each phase was registered before its own runs: Study~2's splits, caches, design tables and analysis scripts were hashed and committed before its first model was trained, with smoke tests on crop-level splits only, and the follow-up after the Study~2 analysis but before any follow-up model or analysis.

\begin{figure*}[t]
\centering
\includegraphics[width=\textwidth]{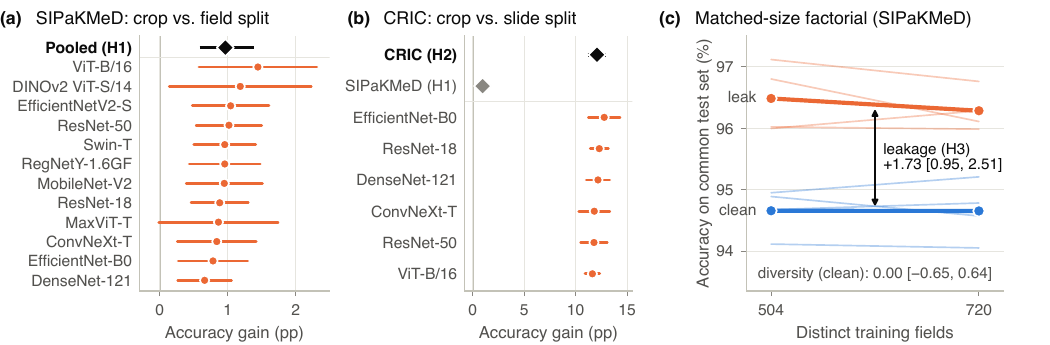}
\caption{Consequence and causal isolation. (a,~b) Crop- minus parent-level accuracy per architecture (mean and 95\% $t$ interval over seeds) and the registered pooled estimates (diamonds). (c) The SIPaKMeD factorial: accuracy on the common test set for leak and clean training sets at two diversity levels (bold: mean of four architectures; faint: each architecture), with the registered leakage and diversity estimates.}
\label{fig:results}
\end{figure*}

\textbf{Protocols and training.} P1 is a stratified crop-level 80:20 split, P2 a parent-disjoint split of equal test size, and P3 additionally merges parents linked by near-duplicates. Twelve ImageNet-pretrained backbones (ResNet-18/50, DenseNet-121, MobileNet-V2, EfficientNet-B0/V2-S, RegNetY-1.6GF, ConvNeXt-T, Swin-T, MaxViT-T, ViT-B/16, DINOv2 ViT-S/14) are fine-tuned with one deterministic recipe (AdamW, $10^{-4}$, batch 64, 20 epochs, resized-crop and flip augmentation) and scored at the final epoch without model selection; per-crop logits and weights are kept.

\textbf{One test per benchmark (H1, H2).} Instead of one test per architecture, the protocol effect is tested once,
\begin{equation}
\mathrm{acc} \sim x + (1+x \mid \mathrm{arch}) + (1 \mid \mathrm{seed}) + (1 \mid \mathrm{split}),
\label{eq:h1}
\end{equation}
where $x=\pm\frac12$ codes P1 vs.\ P2 and \emph{split}, shared by all architectures within a seed, stops twelve evaluations of one test set from counting as twelve~\cite{bates2015lme4,kuznetsova2017lmertest}. A registered residual check selected a two-way cluster bootstrap as the primary inference, with Satterthwaite, Kenward--Roger, binomial-mixed-model and sign-flip tests as sensitivity analyses. Twelve architectures and twenty seeds gave a simulated power of \PowerChosen\ for a 0.8-pp effect on SIPaKMeD; CRIC uses six architectures and ten seeds.

\textbf{Leakage vs.\ diversity (H3).} The factorial (Fig.~\ref{fig:design}b) fixes one test set per seed, half the crops of a random 30\% of multi-crop SIPaKMeD fields (about 527 crops), and varies only the training set, always 220 crops per class. Clean cells draw from other fields; leak cells swap the last fields of the same permutation for the test fields' remaining siblings. Within a diversity level, leak and clean cells thus share crop count, field count and allocation; 720 vs.\ 504 fields gives a 1.43-fold diversity contrast that brackets the natural 1.18. H3 is the leak effect in a mixed model (four architectures, ten seeds, 160 runs); diversity is estimated within clean cells.

\textbf{Diagnostics.} Study~2 registered background transplants (H4) and a gradient-reversal parent adversary (H5)~\cite{ganin2016dann}. The follow-up tests eight hypotheses (F1--F8, \RecFollowTrainRuns\ new models; Fig.~\ref{fig:design}c). The H3 models are re-scored after background removal, illumination flattening, greyscale, low-pass filtering, or Reinhard transfer~\cite{reinhard2001color} of each cell's color statistics to an unseen field of its own class (F2; the target is chosen by label, so this is a controlled diagnostic); the factorial is retrained with masked backgrounds (F3) or strong HED color jitter~\cite{tellez2019quantifying} (F4). Parent recoverability $R$, the within-class accuracy of a logistic field probe on frozen features, is lowered by rank-limited LEACE erasure with a refitted head~\cite{belrose2023leace} (F5), stain normalisation (F6) and a probe-matched linear adversary (F7), each required to remove a quarter of the above-floor $R$ while costing at most 1\pp\ of P2 accuracy; F8 pools the dose--response, and F1 tests a slide-level label prior in CRIC. Grad-CAM~\cite{selvaraju2017gradcam} attribution outside the cell mask is compared descriptively, with a randomised-head check~\cite{adebayo2018sanity}. H1 stands alone; H2--H5 and F1--F8 are separate Holm families~\cite{holm1979}.

\section{Results}
\label{sec:results}

All \RecStudyTwoRuns\ study runs and \RecFollowTasks\ follow-up tasks completed, none diverged, and every result matched its registered split hash. Table~\ref{tab:tests} lists all registered tests.

\subsection{Crop-level splitting inflates both benchmarks}
\label{sec:res-consequence}

On SIPaKMeD, accuracy was \SipPOneAcc\% under P1 and \SipPTwoAcc\% under P2 (H1 $=$ \HoneEst\pp, \HoneBootLo--\HoneBootHi), with all four sensitivity analyses at $p$~\HoneKRp. Every architecture scored higher under P1, by \HoneArchMin--\HoneArchMax\pp\ (Fig.~\ref{fig:results}a): crop-level splitting hides a quarter of the errors. Merging near-duplicate-linked parents changed nothing (P2 $-$ P3: \PtwoPthreeEst\pp, \PtwoPthreeLo\ to \PtwoPthreeHi), so shared parents, not stray duplicates, carry the effect, and the architecture ranking was largely preserved (Kendall $\tau=\KendallTau$). On CRIC, with about 29 mixed-class cells per slide, accuracy fell from \CricPOneAcc\% to \CricPTwoAcc\% (H2 $=$ \HtwoEst\pp; 11.6--12.8\pp\ per architecture; Fig.~\ref{fig:results}b) and balanced accuracy from \CricPOneBacc\% to \CricPTwoBacc\%; a size-matched crop-level control gave the same gap (\CricPrimeEst\pp). That control fixes training size, not slide composition, and the factorial below was run on SIPaKMeD only, so the CRIC gap is consistent with acquisition leakage but not causally decomposed. Our CRIC inputs are $100\times100$ windows around the published nucleus coordinates, so absolute accuracies are not comparable with other CRIC results.

\subsection{On SIPaKMeD, the inflation is leakage, not diversity}
\label{sec:res-isolation}

At fixed crop count, field count and allocation, adding the test fields' siblings raised accuracy on the common test set by \HthreeEst\pp\ (\HthreeLo--\HthreeHi) in every architecture at both diversity levels, while 43\% more fields without leakage did nothing (\DivEst\pp, \DivLo\ to \DivHi; interaction \InteractionEst\pp, $p=\InteractionP$; Fig.~\ref{fig:results}c). The leak replicated on fresh seeds (\HthreeRepEst\pp) and persisted with every background masked in training and testing (F3: \FthreeEst\pp). In an exploratory split it was as large on test crops sharing no pixel with any training crop (\OverlapNo\pp) as on overlapping ones (\OverlapYes\pp): what helps is a seen field, not memorised pixels.

\begin{table}[t]
\caption{All registered tests, in pp (H5, F7, F8: pp per pp of $R$; F1: pp of probability; F6: pp per unit strength), signed so that each hypothesis predicts a positive value. $p$: H1 bootstrap, others Holm. $^\dagger$Significant in the opposite direction.}
\label{tab:tests}
\centering\small\setlength{\tabcolsep}{2.6pt}
\begin{tabular}{@{}l@{\hspace{4pt}}lr@{\hspace{5pt}}cr@{\hspace{3pt}}c@{}}
\toprule
 & Hypothesis & Est. & 95\% CI & $p$ & \\
\midrule
H1 & SIPaKMeD crop vs.\ field & \HoneEst & \HoneBootLo, \HoneBootHi & \HoneBootP & \yes \\
H2 & CRIC crop vs.\ slide & \HtwoEst & \HtwoLo, \HtwoHi & \HtwoPholm & \yes \\
H3 & Leakage, matched size & \HthreeEst & \HthreeLo, \HthreeHi & \HthreePholm & \yes \\
H4 & Own-background route & \HfourEst & \HfourLo, \HfourHi & \HfourPholm & \no \\
H5 & Gap tracks $R$ (GRL) & \HfiveEst & \HfiveLo, \HfiveHi & \HfivePholm & \no \\
\midrule
F1 & CRIC slide-majority pull & \FoneEst & \FoneLo, \FoneHi & \FonePholm & \no$^\dagger$ \\
F2 & color transfer cuts leak & \FtwoEst & \FtwoLo, \FtwoHi & \FtwoPholm & \yes \\
F3 & Leak, no background & \FthreeEst & \FthreeLo, \FthreeHi & \FthreePholm & \yes \\
F4 & HED jitter cuts leak & \FfourEst & \FfourLo, \FfourHi & \FfourPholm & \no \\
F5 & LEACE erasure cuts gap & \FfiveEst & \FfiveLo, \FfiveHi & \FfivePholm & \no \\
F6 & Normalisation cuts gap & \FsixEst & \FsixLo, \FsixHi & \FsixPholm & \no \\
F7 & Gap tracks $R$ (linear) & \FsevenEst & \FsevenLo, \FsevenHi & \FsevenPholm & \no \\
F8 & Gap tracks $R$ (pooled) & \FeightEst & \FeightLo, \FeightHi & \FeightPholm & \no \\
\bottomrule
\end{tabular}
\end{table}

\begin{figure*}[t]
\centering
\includegraphics[width=\textwidth]{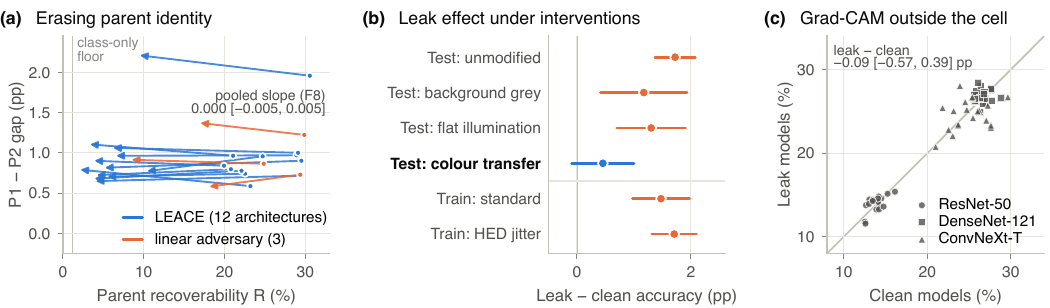}
\caption{Diagnostics. (a) Each arrow takes one architecture from no manipulation to the strongest in-regime LEACE erasure or linear adversary: recoverability falls, the gap does not. (b) Leak minus clean accuracy under test-time interventions (color transfer targets an unseen field of the cell's own class) and training-time color jitter. (c) Grad-CAM attribution outside the cell mask for leak and clean models of the same seed and diversity level; the diagonal marks equality.}
\label{fig:diagnostics}
\end{figure*}

\subsection{Linear decodability and saliency do not track the leak}
\label{sec:res-diagnostics}

Neither registered Study~2 mechanism test located the leaked signal: background transplants (H4) and gradient reversal (H5), which ConvNeXt-T evaded by making parents \emph{more} decodable ($R$ \GrlConvRzero\% to \GrlConvRstar\%), were not supported. Reducing the linearly decodable parent code did not move the gap either (Fig.~\ref{fig:diagnostics}a). Rank-256 LEACE cut $R$ from \LeaceRzero\% to \LeaceRtop\%, removing \LeaceAboveFloorCut\% of its above-floor value (class-only floor \RFloor\%), in all twelve architectures at a \LeaceAccDrop-pp cost in P2 accuracy, and the gap stayed at \LeaceGapZero--\LeaceGapTop\pp\ (F5); the linear adversary passed its check in all three architectures with no change (F7); pooled over \FeightCells\ cells the slope was \FeightEst\ (\FeightLo\ to \FeightHi; F8), so a 20-pp loss of $R$ moves the gap by at most 0.1\pp. Grad-CAM put \CamBgAttribution\% of attribution outside the cell in leak and clean models alike (difference \CamLeakEst\pp, \CamLeakLo\ to \CamLeakHi; Fig.~\ref{fig:diagnostics}c), although its maps passed the randomised-head check (median Spearman $\rho=\CamSpearman$). On CRIC, seen slides pulled minority-class cells \emph{away} from the slide majority (F1: \FoneEst\pp), which argues against a simple slide-majority prior.

Only color transfer moved the leak. Matching each test cell's color statistics to an unseen field of its own class cut the leak from \LeakUnderOriginal\ to \LeakUnderStain\pp\ (F2: \FtwoEst\pp, \FtwoLo--\FtwoHi), whereas flattening illumination (\LeakUnderFlat\pp) or removing the background (\LeakUnderBg\pp), at comparable accuracy cost, did not (Fig.~\ref{fig:diagnostics}b); strong HED jitter in training left it intact (\CarrierColour\ vs.\ \CarrierStandard\pp; F4). Because the transfer is label-conditioned and shifts the test distribution, it implicates field-specific appearance without identifying what the networks learned.

\section{Discussion and conclusion}
\label{sec:discussion}

Crop-level splitting inflates both public cervical cytology benchmarks we trained on. On SIPaKMeD the inflation is leakage: it appears when a test cell's acquisition siblings are in training, not when training merely gains parent diversity. On CRIC, with 29 mixed-class cells per slide instead of four single-class crops per field, the gap is twelve times larger; without a factorial there, it is consistent with, not attributed to, sibling leakage. A benchmark's exposure can be computed from the closed forms of Sec.~\ref{sec:data} before any model is trained. Rankings largely survived on SIPaKMeD, but accuracies under crop- and parent-level splits are not comparable: on CRIC the protocol gap exceeds the spread between architectures.

The mechanism is narrowed, not identified. The advantage survives background masking, does not depend on shared pixels, does not follow a simple slide-majority prior and is unaffected by reducing the linear parent code, while label-conditioned color transfer implicates field-specific appearance; nonlinear or spatially encoded parent information may still carry it. The methodological lesson holds regardless: models whose measured parent code was largely erased, and whose saliency matched that of clean models, still benefited from leakage. Probes establish availability, not use~\cite{elazar2021amnesic,ravichander2021probing}, and neither probes, saliency nor augmentation can stand in for a grouped split.

\textbf{Limitations.} Parents are the deepest released units, not patients: parent-disjoint evaluation removes the leakage measured here, but patient-level grouping could reveal further dependence that these experiments cannot bound. The factorial trains on 1,100 crops rather than 3,239, establishing the leak's existence and direction more firmly than an exact decomposition of H1. One training recipe was used, erasure acts on linearly readable pooled features, and the follow-up, registered after the Study~2 results, is confirmatory only for its own questions.

\textbf{Recommendations.} Releases should label every observation with its parent, and its patient where permitted; evaluations should split at the deepest released unit, publish hashed fold manifests and report the fraction of test observations with a training sibling. Code, provenance maps, split manifests, run records and a risk calculator are at \url{https://github.com/optimisticexquisite/cytofair}.

\section{Compliance with ethical standards}
\label{sec:ethics}

This research was conducted retrospectively using de-identified human cytology images made available in open access by the creators of SIPaKMeD, Cx22, CRIC and Herlev. No new data were collected, and ethical approval was not required for this secondary analysis of public data.

\section{Acknowledgments}
\label{sec:acknowledgments}

No funding was received for conducting this study. The authors have no relevant financial or non-financial interests to disclose. Claude (Anthropic), used through Claude Code, assisted in writing and debugging the experiment and analysis code, orchestrating the training runs, producing the figures, and drafting and editing the text of all sections. The authors reviewed all code, results, references and text, and take full responsibility for the content.

\bibliographystyle{IEEEbib}
\bibliography{refs}

@inproceedings{plissiti2018sipakmed,
  author    = {Marina E. Plissiti and P. Dimitrakopoulos and G. Sfikas and Christophoros Nikou and O. Krikoni and A. Charchanti},
  title     = {{SIPaKMeD}: A New Dataset for Feature and Image Based Classification of Normal and Pathological Cervical Cells in {Pap} Smear Images},
  booktitle = {Proc. IEEE Int. Conf. Image Process. (ICIP)},
  year      = {2018},
  pages     = {3144--3148},
  doi       = {10.1109/ICIP.2018.8451588}
}

@article{liu2022cx22,
  author  = {Guangqi Liu and Qinghai Ding and Haibo Luo and Min Sha and Xiang Li and Moran Ju},
  title   = {{Cx22}: A New Publicly Available Dataset for Deep Learning-Based Segmentation of Cervical Cytology Images},
  journal = {Comput. Biol. Med.},
  volume  = {150},
  pages   = {106194},
  year    = {2022},
  doi     = {10.1016/j.compbiomed.2022.106194}
}

@article{rezende2021cric,
  author  = {Mariana T. Rezende and others},
  title   = {{CRIC} Searchable Image Database as a Public Platform for Conventional {Pap} Smear Cytology Data},
  journal = {Scientific Data},
  volume  = {8},
  number  = {1},
  pages   = {151},
  year    = {2021},
  doi     = {10.1038/s41597-021-00933-8}
}

@article{kapoor2023leakage,
  author  = {Sayash Kapoor and Arvind Narayanan},
  title   = {Leakage and the Reproducibility Crisis in Machine-Learning-Based Science},
  journal = {Patterns},
  volume  = {4},
  number  = {9},
  pages   = {100804},
  year    = {2023},
  doi     = {10.1016/j.patter.2023.100804}
}

@article{geirhos2020shortcut,
  author  = {Robert Geirhos and J{\"o}rn-Henrik Jacobsen and Claudio Michaelis and Richard Zemel and Wieland Brendel and Matthias Bethge and Felix A. Wichmann},
  title   = {Shortcut Learning in Deep Neural Networks},
  journal = {Nat. Mach. Intell.},
  volume  = {2},
  number  = {11},
  pages   = {665--673},
  year    = {2020},
  doi     = {10.1038/s42256-020-00257-z}
}

@article{varoquaux2022failures,
  author  = {Ga{\"e}l Varoquaux and Veronika Cheplygina},
  title   = {Machine Learning for Medical Imaging: Methodological Failures and Recommendations for the Future},
  journal = {npj Digital Medicine},
  volume  = {5},
  pages   = {48},
  year    = {2022},
  doi     = {10.1038/s41746-022-00592-y}
}

@inproceedings{jantzen2005pap,
  author    = {Jan Jantzen and Jonas Norup and Georgios Dounias and Beth Bjerregaard},
  title     = {Pap-Smear Benchmark Data for Pattern Classification},
  booktitle = {Proc. Nature Inspired Smart Inf. Syst. (NiSIS)},
  year      = {2005},
  pages     = {1--9},
  address   = {Albufeira, Portugal},
  publisher = {NiSIS}
}

@article{bates2015lme4,
  author  = {Douglas Bates and Martin M{\"a}chler and Ben Bolker and Steve Walker},
  title   = {Fitting Linear Mixed-Effects Models Using {lme4}},
  journal = {J. Stat. Softw.},
  volume  = {67},
  number  = {1},
  pages   = {1--48},
  year    = {2015},
  doi     = {10.18637/jss.v067.i01}
}

@article{kuznetsova2017lmertest,
  author  = {Alexandra Kuznetsova and Per B. Brockhoff and Rune H. B. Christensen},
  title   = {{lmerTest} Package: Tests in Linear Mixed Effects Models},
  journal = {J. Stat. Softw.},
  volume  = {82},
  number  = {13},
  pages   = {1--26},
  year    = {2017},
  doi     = {10.18637/jss.v082.i13}
}

@article{ganin2016dann,
  author  = {Yaroslav Ganin and Evgeniya Ustinova and Hana Ajakan and Pascal Germain and Hugo Larochelle and Fran{\c{c}}ois Laviolette and Mario Marchand and Victor Lempitsky},
  title   = {Domain-Adversarial Training of Neural Networks},
  journal = {J. Mach. Learn. Res.},
  volume  = {17},
  number  = {59},
  pages   = {1--35},
  year    = {2016}
}

@article{reinhard2001color,
  author  = {Erik Reinhard and Michael Ashikhmin and Bruce Gooch and Peter Shirley},
  title   = {Color Transfer between Images},
  journal = {IEEE Comput. Graph. Appl.},
  volume  = {21},
  number  = {5},
  pages   = {34--41},
  year    = {2001},
  doi     = {10.1109/38.946629}
}

@article{tellez2019quantifying,
  author  = {Tellez, David and Litjens, Geert and B{\'a}ndi, P{\'e}ter and Bulten, Wouter and Bokhorst, John-Melle and Ciompi, Francesco and van der Laak, Jeroen},
  title   = {Quantifying the Effects of Data Augmentation and Stain Color Normalization in Convolutional Neural Networks for Computational Pathology},
  journal = {Med. Image Anal.},
  volume  = {58},
  pages   = {101544},
  year    = {2019}
}

@inproceedings{belrose2023leace,
  author    = {Belrose, Nora and Schneider-Joseph, David and Ravfogel, Shauli and Cotterell, Ryan and Raff, Edward and Biderman, Stella},
  title     = {{LEACE}: Perfect Linear Concept Erasure in Closed Form},
  booktitle = {Adv. Neural Inf. Process. Syst. (NeurIPS)},
  volume    = {36},
  year      = {2023}
}

@inproceedings{selvaraju2017gradcam,
  author    = {Selvaraju, Ramprasaath R. and Cogswell, Michael and Das, Abhishek and Vedantam, Ramakrishna and Parikh, Devi and Batra, Dhruv},
  title     = {Grad-{CAM}: Visual Explanations from Deep Networks via Gradient-Based Localization},
  booktitle = {Proc. IEEE Int. Conf. Comput. Vis. (ICCV)},
  pages     = {618--626},
  year      = {2017}
}

@inproceedings{adebayo2018sanity,
  author    = {Adebayo, Julius and Gilmer, Justin and Muelly, Michael and Goodfellow, Ian and Hardt, Moritz and Kim, Been},
  title     = {Sanity Checks for Saliency Maps},
  booktitle = {Adv. Neural Inf. Process. Syst. (NeurIPS)},
  volume    = {31},
  year      = {2018}
}

@article{holm1979,
  author  = {Sture Holm},
  title   = {A Simple Sequentially Rejective Multiple Test Procedure},
  journal = {Scand. J. Stat.},
  volume  = {6},
  number  = {2},
  pages   = {65--70},
  year    = {1979}
}

@article{elazar2021amnesic,
  author  = {Elazar, Yanai and Ravfogel, Shauli and Jacovi, Alon and Goldberg, Yoav},
  title   = {Amnesic Probing: Behavioral Explanation with Amnesic Counterfactuals},
  journal = {Trans. Assoc. Comput. Linguist.},
  volume  = {9},
  pages   = {160--175},
  year    = {2021}
}

@inproceedings{ravichander2021probing,
  author    = {Ravichander, Abhilasha and Belinkov, Yonatan and Hovy, Eduard},
  title     = {Probing the Probing Paradigm: Does Probing Accuracy Entail Task Relevance?},
  booktitle = {Proc. Conf. European Chapter Assoc. Comput. Linguist. (EACL)},
  pages     = {3363--3377},
  year      = {2021}
}

@inproceedings{jimenez2023shortcuts,
  author    = {Amelia Jim{\'e}nez-S{\'a}nchez and Dovile Juodelyte and Bethany Chamberlain and Veronika Cheplygina},
  title     = {Detecting Shortcuts in Medical Images -- A Case Study in Chest {X}-Rays},
  booktitle = {Proc. IEEE Int. Symp. Biomedical Imaging (ISBI)},
  pages     = {1--5},
  year      = {2023},
  doi       = {10.1109/ISBI53787.2023.10230572}
}

@inproceedings{lin2024shortcut,
  author    = {Manxi Lin and Nina Weng and Kamil Mikolaj and Zahra Bashir and Morten Bo S{\o}ndergaard Svendsen and Martin Tolsgaard and Anders Nymark Christensen and Aasa Feragen},
  title     = {Shortcut Learning in Medical Image Segmentation},
  booktitle = {Proc. Med. Image Comput. Comput. Assist. Interv. (MICCAI)},
  pages     = {623--633},
  year      = {2024},
  doi       = {10.1007/978-3-031-72111-3_59}
}

@article{germani2026aggregation,
  author  = {Elodie Germani and Fatima Zeineddine and Charbel Mourad and Shadi Albarqouni},
  title   = {When Multi-Institutional Dataset Aggregation Masks Shortcut-Like Behavior in Foundation Models for Medical Imaging},
  journal = {Med. Image Anal.},
  volume  = {114},
  pages   = {104258},
  year    = {2026},
  doi     = {10.1016/j.media.2026.104258}
}

@misc{vangala2026scores,
  author       = {Bhanu Prakash Vangala and Sowmya Guda and Latha Peddi and Navya Vangala},
  title        = {Scores That Hold, Benchmarks That Leak: Measuring Dataset Contamination in Public Brain-Tumor {MRI} Classification},
  howpublished = {arXiv:2610.00421},
  year         = {2026}
}

@misc{obeten2026kidney,
  author       = {Okoi Michael Obeten and Abas Aliu and Omowumi Hafsat Aliu and Ahmed Jimoh and Aliyu Zibril and Uchenna Christiana Ezeanya},
  title        = {Data Leakage Inflates Reported Accuracy of Deep Learning for Kidney Stone Detection on {CT}: A Leakage-Free, Calibrated and Uncertainty-Aware Reassessment across Three Centers},
  howpublished = {SSRN preprint 7117201},
  year         = {2026}
}

@inproceedings{bussola2021tiles,
  author    = {Nicole Bussola and Alessia Marcolini and Valerio Maggio and Giuseppe Jurman and Cesare Furlanello},
  title     = {{AI} Slipping on Tiles: Data Leakage in Digital Pathology},
  booktitle = {Proc. ICPR Int. Workshops and Challenges},
  series    = {LNCS},
  volume    = {12661},
  pages     = {167--182},
  year      = {2021},
  doi       = {10.1007/978-3-030-68763-2_13}
}

\end{document}